\documentclass[letterpaper, 10 pt, conference]{ieeeconf}  

\IEEEoverridecommandlockouts                              
\usepackage{graphics} 
\usepackage{epsfig} 
\usepackage{mathptmx} 
\usepackage{times} 
\usepackage{amsmath, amsthm, amssymb, amsfonts}
\usepackage{float}

\usepackage{eucal}

\usepackage{tabularx}
\usepackage{booktabs}
\usepackage{multirow}
\usepackage{mathtools}

\usepackage{siunitx}
\usepackage{xcolor}
\usepackage[normalem]{ulem}

\usepackage{balance}
\usepackage[nospace]{cite}
\usepackage{url}
\usepackage{verbatim}

\title{\LARGE \bf
Modeling, Reduction, and Control of a Helically Actuated Inertial\\Soft Robotic Arm via the Koopman Operator}

\author{David A. Haggerty$^{1,2,3}$, Michael J. Banks$^{1,3}$, Patrick C. Curtis$^{1}$, Igor Mezi\'c$^{1}$, and Elliot W. Hawkes$^{1}$
\thanks{*This work was supported in part by the National Science Foundation grant no. 1935327. }
\thanks{$^{1}$Authors are with the Department of Mechanical Engineering, University of California, Santa Barbara,
        Santa Barbara, CA, 93106.}%
\thanks{$^{2}$Corresponding author, 
        {\tt\small davidhaggerty@ucsb.edu}.}%
\thanks{$^{3}$David A. Haggerty and Michael J. Banks are co-first authors.}
}

\begin{document}

\maketitle
\thispagestyle{empty}
\pagestyle{empty}

\begin{abstract}

Soft robots promise improved safety and capability over rigid robots when deployed in complex, delicate, and dynamic environments. However the infinite degrees of freedom and highly nonlinear dynamics of these systems severely complicate their modeling and control. As a step toward addressing this open challenge, we apply the data-driven, Hankel Dynamic Mode Decomposition (HDMD) with time delay observables to the model identification of a highly inertial, helical soft robotic arm with a high number of underactuated degrees of freedom. The resulting model is linear and hence amenable to control via a Linear Quadratic Regulator (LQR). Using our test bed device, a dynamic, lightweight pneumatic fabric arm with an inertial mass at the tip, we show that the combination of HDMD and LQR allows us to command our robot to achieve arbitrary poses using only open loop control. We further show that Koopman spectral analysis gives us a dimensionally reduced basis of modes which decreases computational complexity without sacrificing predictive power.



\end{abstract}

\section{Introduction}
\label{sec:intro}

While soft robotics has garnered significant attention in the past decade and grown into a standalone research topic, one of the prevailing challenges the field faces is the problem of modeling and control. The high degrees of freedom, material non-linearity, underactuation, and inherent hysteresis of many of these technologies has precluded the development of closed form, dynamic models that easily lend themselves to traditional control strategies \cite{rus2015design,laschi2016soft,trivedi2008soft}. Instead, a variety of methods have been introduced in an attempt to address this challenge.

\begin{figure}[t]
    \centering
    \includegraphics[width=0.95\columnwidth]{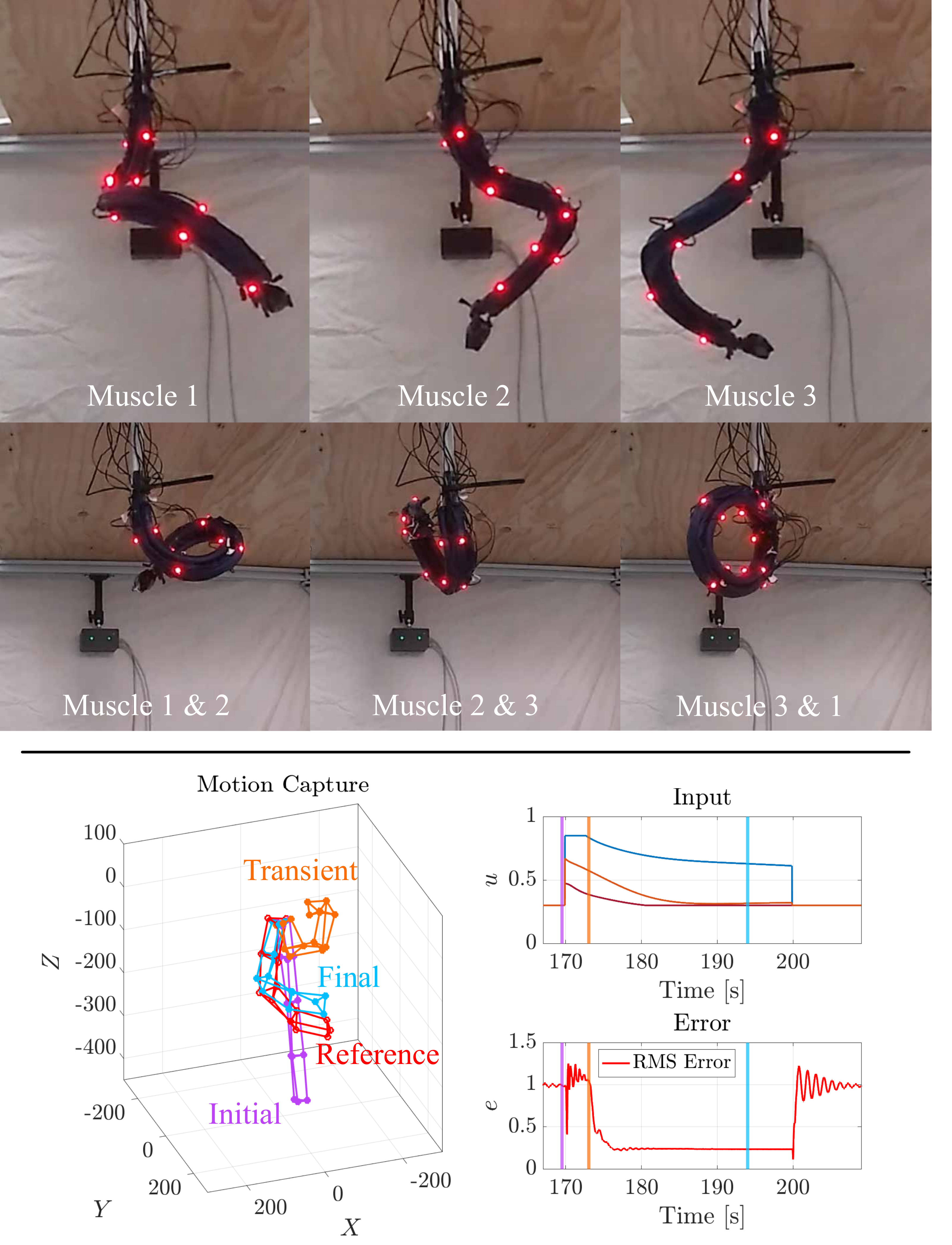}
    \caption{\emph{Top:} The steady state pose of the helically actuated, inertial soft arm with an input to individual and pairs of lengthwise artificial muscles. \emph{Bottom:} With a linear model constructed via data-driven, Koopman operator theoretic Hankel Dynamic Mode Decomposition (HDMD) and reduced in dimensionality to $n=35$ using Koopman mode analysis, we use the Linear Quadratic Regulator (LQR) optimal control algorithm to control the robot from an initial position (purple) through a dynamic transient pose (orange) to its final pose (blue), closely aligned with the target position (red). Plots of the commanded inputs and RMS error are shown, at right.}
    \label{fig:introFig}
     \vspace{-5mm}
\end{figure}

A majority of investigations to this end have relied on simplifying assumptions, such as the (piecewise) constant curvature ((P)CC) approach, as found in for example \cite{hannan2003kinematics,trivedi2008geometrically,jones2006practical,wang2020dexterous,greer2018soft,selvaggio2020obstacle}. Most of these approaches focus on developing mappings from the actuator space (actuator pressure, tendon tension) to the configuration space (curvature, arc length) and finally to the task space (euclidean position and angle). While this approach is macroscopically effective at predicting general deformations, it fails to adequately capture the time evolution of relevant quantities. As such, others have sought to build closed-form dynamic models, compensated with controllers based on feedback linearization \cite{falkenhahn2015model}, sliding mode control \cite{kapadia2010model}, and domain restriction \cite{marchese2016dynamics}. However, due to the infinite degrees of freedom of these systems, closed form models are inherently inaccurate. Moreover, the intrinsic non-linearity often ensures control systems be domain restricted or themselves non-linear.

Owing to these difficulties, a number of groups have turned to data-driven approaches for producing linear system representations. Numerous attempts to apply various machine learning methods span the last decade, as in \cite{engel2006learning, braganza2007neural, thuruthel2017learning,della2020data,della2020model}, for example. In each of these investigations, the overarching aim has been to fit a high dimensional linear operator to the input-output dynamics of a soft robot arm. While they have shown much higher efficacy than many of the closed-form approaches above, they yet require immense amounts of data to converge, and their models don't necessarily intuit any physical characteristics of the system. As such, they often produce very large linear systems with limited domains of applicability. To overcome these limitations, a relatively new attention paid to a century-old approach in dynamical systems theory has opened new avenues to accurate, dynamically relevant models. This approach, Koopman Operator Theory (KOT) \cite{Koopman1931}, has been shown to be more effective than other data-driven methods for soft robotic modeling and control \cite{bruder2019nonlinear}. However, KOT applied to soft robotics is still in its infancy and has yet to be utilized to its full theoretical potential. 

This work aims to advance the state-of-the-art in KOT applied to soft robotics through the analysis of Koopman modes. We do so by implementing HDMD on a tip-loaded, inertial soft robotic arm exhibiting both bending and twisting (Fig. \ref{fig:introFig}). Using the spatial positions of $15$ motion tracking points and their time delays as observables, HDMD captures the fundamental physics of our system \cite{arbabi2017ergodic}. KOT enables us to weigh the relative importance of each of the system's fundamental modes to the dynamics we are interested in. We then project the resulting model onto a reduced basis of the most important Koopman modes. This approach enables us to substantially reduce the order of the model without significant loss of controllability. Notably, we do so with no pre-optimization of observables or extensive data postprocessing, and with training data on the order of $10^4$ samples.

What follows is a description of our soft robotic arm, an introduction to KOT and the details of our modeling approach, our experimental setup and data acquisition methods, our results, and a discussion of open questions to be addressed in future work.

\section{Soft Robotic Arm Design}
\label{sec:arm}

Due to the difficulty of the problem at hand, one approach is to simplify the testbed to simplify the modeling. Instead, to understand the limitations of the KOT based modeling, we sought to produce a difficult-to-analyze soft arm. This manifested in the concurrent objectives of: fast response times, highly non-linear deformation, and a highly inertial and underactuated system.

We created a pneumatic system capable of generating low-latency, agile actuation over a wide range of inputs. This design was fabricated with lightweight 50\,micron thick silicone-impregnated ripstop nylon (sil-nylon), and actuated via three lengthwise fabric artificial muscles of the same material, as presented in \cite{naclerio2020simple}. The main body was fabricated to a 2\,cm diameter using a silicone adhesive (Smooth-On Sil-poxy) with the fiber reinforcement aligned axially. The three 1.25\,cm diameter muscles were similarly constructed, however their fiber weave was oriented with a $45^\circ$ offset with respect to the main body's axial direction. The mass of the arm with three muscles is a mere 12\,g. The input valves that supply the air to the muscles were chosen to enable at least 60 L/min of flow at 200\,kPa to ensure high power input. Additionally, the exhaust valves through which the air leaves empty to vacuum to increase the speed of the robot.


To achieve a complicated, non-linear actuation pattern, the muscles were axially aligned on the body with a slight offset, varying among the muscles, to produce different helical deformations from each (Fig. \ref{fig:introFig}, \emph{Top}) as described in \cite{blumenschein2018helical}. To achieve extreme curvature, the main body was fixtured to a workbench and the muscles were affixed under pretension. 

Finally, to create a highly inertial system with deformations not directly controlled by the muscles, a 40\,g mass was adhered to the tip of the robot.


\section{Control System Modeling via Koopman Operator Theory}
\label{sec:model}

The standard method of representing dynamical systems involves defining a state space $M$ with states $x\in M$ that evolve according to the discrete-time dynamical system 
\begin{equation}
    x^+=T(x).
    \label{eqn:dynamicalMap}
\end{equation}
Here $T$ is the possibly nonlinear state transition function $T:M\to M$.

The non-linearity of soft robot dynamics limits the availability of suitable state-space control algorithms. We instead turn to an operator-theoretic perspective of dynamics of observables \cite{Koopman1931}. Observables are real-valued functions defined on the state space $f:M\to\mathbb{R}$. The set of all possible observables forms a vector space that is usually infinite dimensional. The Koopman operator $\mathcal{K}$ is defined by
\begin{equation*}
    \mathcal{K}f \coloneqq f\circ T.
\end{equation*}
This operator describes the evolution of observables as the states move along orbits dictated by \eqref{eqn:dynamicalMap}. Even though the underlying state space system is nonlinear, the operator $\mathcal{K}$ is linear. The process of approximating this operator with a finite dimensional matrix is described in Section \ref{subsec:koopmanDMD}.

We are interested in describing the Koopman operator for systems of the form $x^+=T(x,u)$ where $u\in \mathcal{U}$ are user specified inputs. We follow the process outlined in \cite{Korda2018koopmpc} to build this generalization. The first step is to define the space of all input sequences $l(\mathcal{U})=\{(u_i)_{i=0}^{\infty}|u_i\in \mathcal{U}\}$ where $\mathcal{U}$ is the set of admissible inputs. The discrete-time dynamics $T$ now act on the extended state space $S=M\times l(\mathcal{U})$.
Given observables $g:S\to\mathbb{R}$, we now define the corresponding Koopman operator
\begin{equation}
    (\mathcal{K}g)(x,(u_i)_{i=0}^{\infty})\coloneqq g(T(x,u_0),(u_i)_{i=0}^{\infty})
    \label{eqn:koopOpDef}
\end{equation}
We seek a finite dimensional linear input/output system which approximates the action of $\mathcal{K}$ on a finite set of chosen observables.

\subsection{Approximation of Koopman Operators for Control Systems}
\label{subsec:koopmanDMD}
\begin{figure}[tb]
    \centering
    \includegraphics[width=0.7\columnwidth]{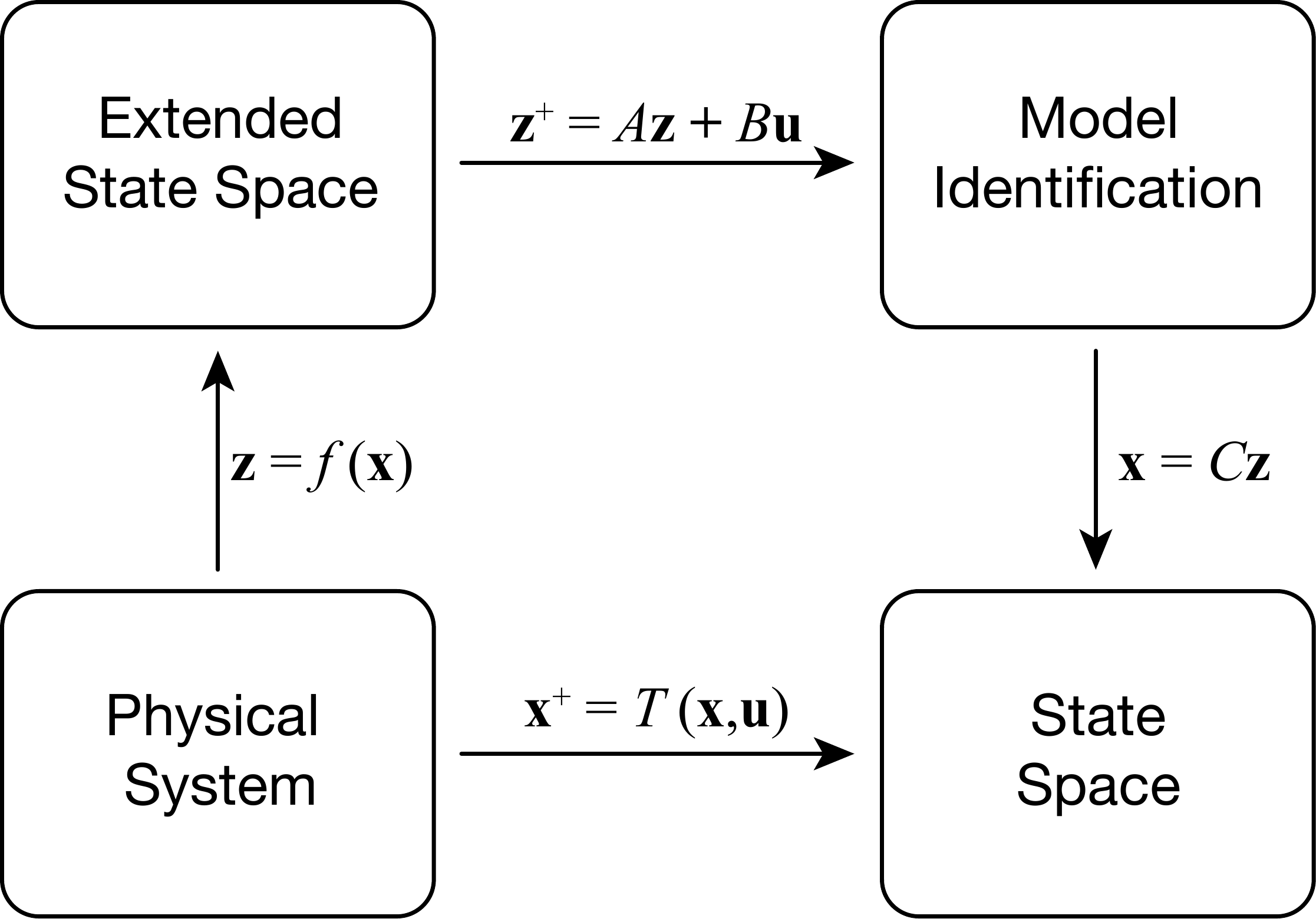}
    \caption{Description of Koopman lifting process and reduction to a state space representation. If the (possibly nonlinear) map $T(\textbf{x},\textbf{u})$ is known, a state space representation can be immediately developed. If not, a linear evolution of observables $\textbf{z}$ can be projected onto the state space after learning the relationship $\textbf{z}^+ = A\textbf{z}+B\textbf{u}$.}
    \label{fig:koopTheory}
    \vspace{-5mm}
\end{figure}

The Koopman operator in its fully infinite dimensional form is not practically realizable, so we seek a finite dimensional approximation. Under certain conditions, the Hankel Dynamic Mode Decomposition (HDMD) \cite{arbabi2017ergodic, arbabi2017computation} provably converges to the Koopman operator in the limit of infinitely many observables and data snapshots \cite{Korda2017converge}. The practical considerations behind our choices of observables and generation of training data are discussed in \ref{subsec:koopPred}. The following exposition on the HDMD algorithm is closely based on \cite{Korda2018koopmpc}, and is shown in Fig \ref{fig:koopTheory}.

Given $K$ measurements of the system $x^+_j=T(x_j,u_j)$, we build the following data matrices:
\begin{eqnarray}
    X\coloneqq[x_1 \ ... \ x_K], & X^+\coloneqq[x_1^+ \ ... \ x_K^+], & U\coloneqq[u_1 \ ... \ u_K].
\end{eqnarray}
We then choose a vector of $m$ observables
\begin{equation}
    \mathbf{f}(x)=[f_1(x) \ ... \ f_m(x)]^T.
\end{equation}
Next, we build the lifted data matrices
\begin{eqnarray}
    X_{\texttt{lift}}\coloneqq[\mathbf{f}(x_1) \ ... \ \mathbf{f}(x_K)], & X^+_{\texttt{lift}}\coloneqq[\mathbf{f}(x_1^+) \ ... \ \mathbf{f}(x_K^+)].
\end{eqnarray}
We seek to approximate the action of the extended Koopman operator \eqref{eqn:koopOpDef} as follows:
\begin{equation}
    X^+_{\texttt{lift}}=AX_{\texttt{lift}}+BU
\end{equation}
In order to approximate $A$ and $B$, we recast this equation as a minimization problem
\begin{equation}
    \min_{A,B}\|X^+_{\texttt{lift}}-AX_{\texttt{lift}}-BU\|_F
\end{equation}
which has the solution
\begin{equation}
\left[A \ B\right]=X^+_{\texttt{lift}}\left(\left[\begin{matrix} X_{\texttt{lift}} \\ U \end{matrix}\right]\right)^{\dagger}
\label{eqn:ABdefinition}
\end{equation}
where $\dagger$ is the Moore-Penrose pseudoinverse.
The $A$ and $B$ matrices form a dynamical system relevant not to states in the state space but to an extended set of states formed by the vector of observables $z=\bold{f}(x)$. The resulting system is
\begin{equation}
    z^+=Az+Bu.
    \label{eqn:liftedSystem}
\end{equation}
We are often interested in the spectral properties of the Koopman operator because they give us physical information about the multiple coupled time-dependent processes inherent to our system. HDMD can be used to approximate the discrete part of this spectrum \cite{Korda2017converge}. We seek the triplet $(\lambda_i,\phi_i(x),\bold{v}_i)$ of Koopman eigenvalues, eigenfunctions, and modes, respectively. The eigenvalues and Koopman modes are simply the eigenvalues and eigenvectors of the HDMD matrix $A$. The resulting modes also form a convenient basis onto which we can project our dynamics as demonstrated in Figs \ref{fig:staticLQRpose} and \ref{fig:transientLQRpose}. Computation of the eigenfunctions requires $\bold{w}_i$ which are the eigenvectors of the conjugate transpose of $A$. After these are normalized so that $\langle\bold{v}_i,\bold{w}_j\rangle=\delta_{ij}$, the eigenfunctions are given by the complex inner product $\phi_i(x)=\langle x,\bold{w}_i\rangle$.

\subsection{Koopman-based Optimal Control}
\label{subsec:koopmanLQR}

HDMD gives us the model
\begin{eqnarray}
z^+ & = & Az+Bu \\
x   & = & Cz.
\end{eqnarray}
This model is simple enough that we can apply well known optimal control methods such as the Ricatti-equation based Linear Quadratic Regulator \cite{anderson2007optimal}. We define the cost function
\begin{equation}
    J = \sum_{i=1}^K\left[(x_i-x_{\texttt{ref}})^TQ(x_i-x_{\texttt{ref}})+u_i^TRu_i\right]
    \label{eqn:costFunc}
\end{equation}
where $x_{\texttt{ref}}$ is the desired position and $Q$ and $R$ are diagonal state and input penalty matrices, respectively. The minimizing solution to the cost function $J$ is given by the discrete time algebraic Riccati equation. The resulting gain matrix $K$ is used to develop the control law defined by

\begin{equation}
\begin{aligned}
u_i &= -K(z_i-z_{\texttt{ref}}). \\
z_{i+1} &= Az_i + Bu_i \\
x_i &= Cz_{i+1}.
\end{aligned}
\label{eqn:optControl}
\end{equation}
This signal is the optimal stabilizing solution taking the present initial state to the desired state, $x_{\texttt{ref}}$.


\section{Experimental Setup and Methods}
\label{sec:methods}

\subsection{Setup and Training}
\label{subsec:setup}

To apply the approach presented in Sec. \ref{sec:model} to our robot described in Sec. \ref{sec:arm}, we built a $1.8\,\text{m} \times 1.8\,\text{m} \times 1.5\,\text{m}$ frame using T-slotted aluminum, with the top face 2/3 covered with plywood to support our robot and driving circuitry. A hole was cut into the plywood through which a rigid pipe extension was passed, that included through-holes for the pneumatic tubing. The robot body was affixed to this pipe extension, and tubing routed to the muscles. The pressurization of the muscles is controlled by six Clippard DV-2M-12 proportional valves, with three each for input and exhaust, one input-exhaust pair attached to each muscle. The main body is held at a constant pressure of 100\,kPa throughout testing, while the muscles are each controlled in a range of 0-200\,kPa. 

\begin{figure}[htb]
\centering
    
    \includegraphics[width=0.75\columnwidth]{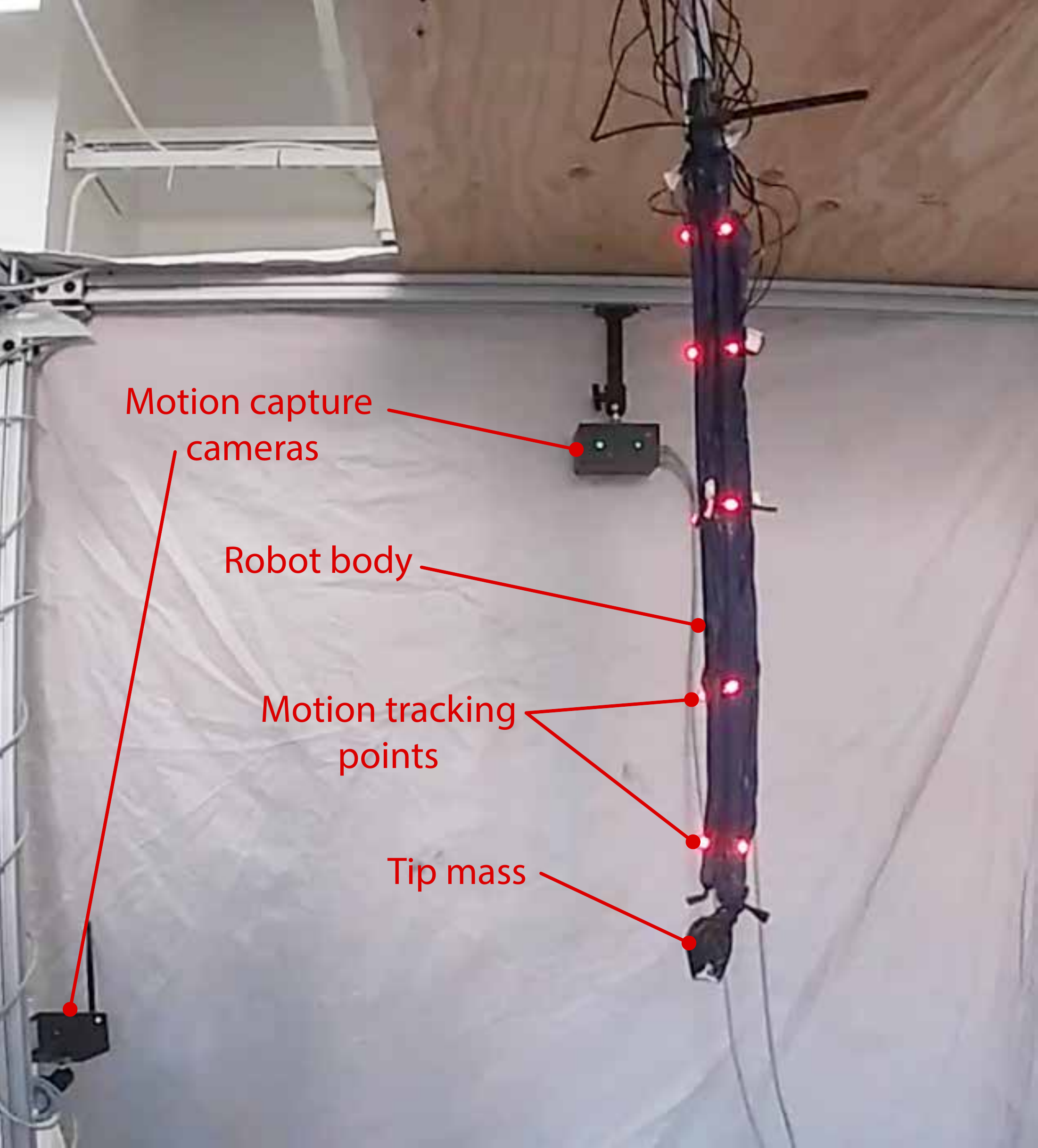}
    \caption{Experimental setup, with the robot in an unactuated state.}
    \label{fig:blockSetUp}
    \vspace{-5mm}
\end{figure}
In order to train our model, we produced inputs that would allow the robot to explore the entire space of configurations that are relevant to our control scenarios. The training inputs were required to be within the bounds $u \in [0.3, 0.85]$ (values given as a percentage of duty cycle), which is the active region for the valves. 
In the first of two training regimes, 150 randomly generated Gaussians 
were superimposed to create a smoothly varying signal that was sufficiently random to guarantee that the robot would explore the entire state space slowly and without overshoot. In the second regime, step inputs of random height were commanded to produce massive overshoot and settling to multiple input-modified equilibria. 
The exhaust signal, $v$, was defined by $1-u$. 
These inputs were then deployed to a Raspberry Pi Model 4B. A trigger signal was also defined to synchronize the Raspberry Pi and our motion capture system.   

Information about the position and shape of the manipulator is gathered via motion capture (PhaseSpace Impulse X2E) with fifteen LED trackers and eight cameras. Five LEDs are attached along the axis of each muscle. 
This entire process is described in Fig. \ref{fig:blockSetUp}.

After 30 minutes of training data were acquired, they were segmented into training sets and verification sets. The model was trained on half the data, while the other half was used for reconstruction and optimal control objectives.

    

\subsection{Verification}
\label{subsec:verification}

After training our model, we sought to apply it towards commanding the robot from the origin to a static pose. As described in Sec. \ref{subsec:setup}, we extracted states the robot achieved from the verification set. 
These states were then incorporated into the optimal control scheme provided in \eqref{eqn:optControl} at a number of different model reductions, and the inputs generated were supplied to the robot as described above. The results for these tests are shown in Sec. \ref{sec:results}.



\section{Results}
\label{sec:results}

Here we present the results of our application of the modeling described in Sec. \ref{sec:model} to the soft robot arm introduced in Sec. \ref{sec:arm}. We first discuss how our model predicts the dynamics of our system with a variety of observable choices, and then continue with the results of our open loop optimal control efforts to achieve two poses. 

\subsection{Koopman for Prediction}
\label{subsec:koopPred}

We perform a convergence study on the reconstruction power of our Koopman models as a function of the number of snapshots for a range of observables. This process allowed us to develop a dictionary of observables suitable for our system. 
Given a particular choice of observables and number of training samples, we build the corresponding linear input-output system with $A$, $B$, and $C$ matrices. This linear model is applied to $N=27000$ samples of verification data. These particular samples are not included in the training data in order to give us a fair evaluation of the predictive power of our models. The linear system produced via \eqref{eqn:ABdefinition} and \eqref{eqn:liftedSystem} evaluate the evolution of these initial conditions over a single time step. The single-step reconstruction error is given by
\begin{equation*}
    e_i = \frac{\|x_{i}^{+, \ \texttt{predict}} - x_{i}^{+, \ \texttt{actual}}\|_F}{\|x_{i}^{+, \ \texttt{actual}} - x_{i}\|_F}.
\end{equation*}
where $x_{i}^{+, \ \texttt{actual}}$ is the evolution of $x_i$ measured by the motion capture system and $x_{i}^{+, \ \texttt{predict}}$ is the evolution predicted by the HDMD model. We use the root mean square (RMS) of the individual $e_i$ errors to score our model:
\begin{equation*}
    e_{\texttt{RMS}}=\sqrt{\frac{1}{N}\sum_{i=1}^Ne_i^2}.
\end{equation*}

\subsubsection{Choice of Observables}
\begin{figure}[tb]
    \centering
    \includegraphics[width=1\columnwidth]{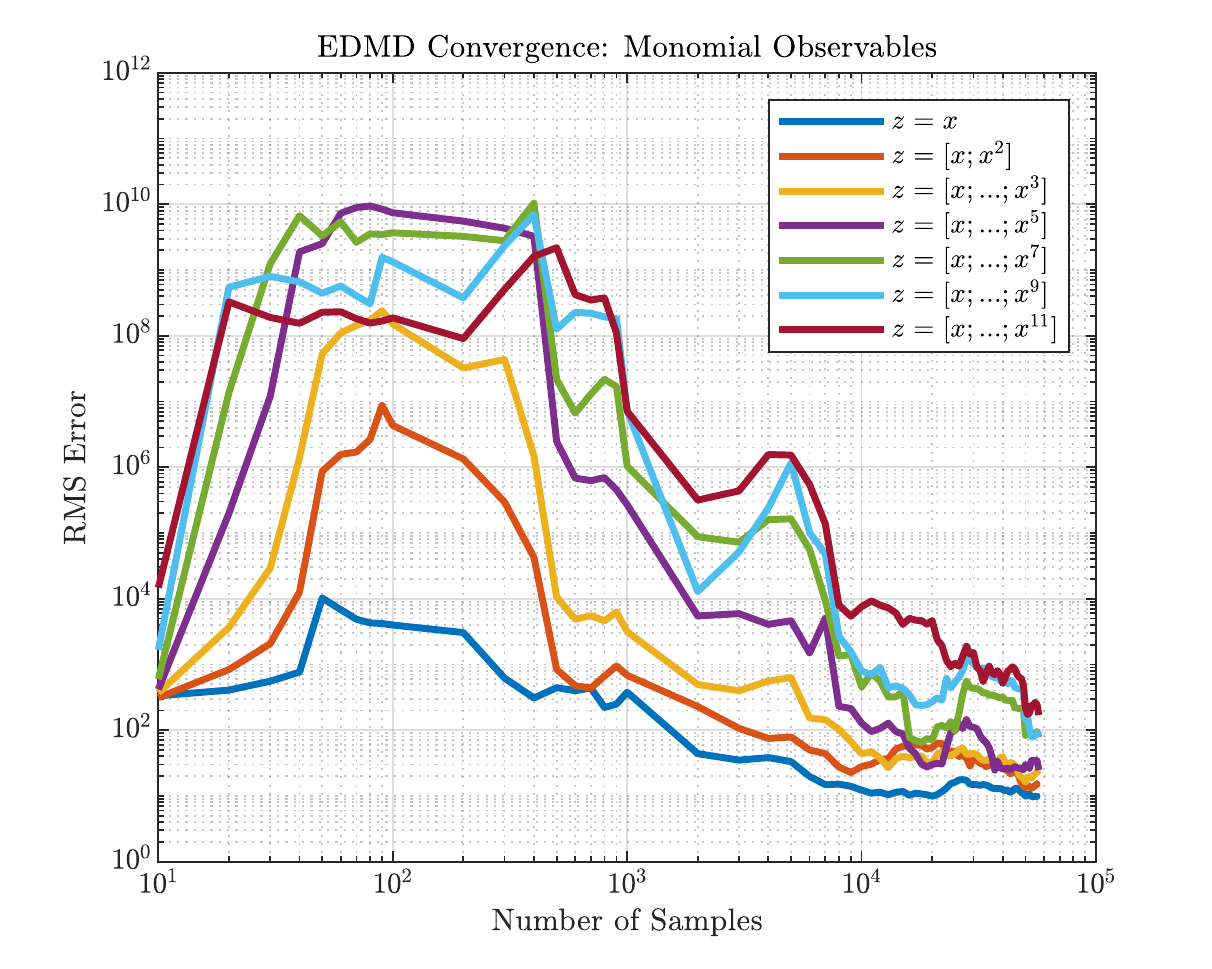}
    \vspace{-5mm}
    \caption{The error $e_{\texttt{RMS}}$ vs number of time samples for a range of monomial observables. Here $x$ indicates a column vector of the positions of all $15$ motion tracker points. The extended state $z=[x;...;x^i]$ is formed by stacking the element-wise powers of $x$ from $x^1$ up to and including $x^i$. We see a decrease in predictive power as the order of monomials increases.}
    \label{fig:conv_mono}
    \vspace{-5mm}
\end{figure}

We tested two different choices for observables. First, we used 
the set of monomials ranging from order 1-4, as described in \cite{bruder2019modeling}, as well as additional monomials up to order 11. 
We found that these basis functions performed poorly for our highly inertial, non-linear system (Fig. \ref{fig:conv_mono}). As can be seen, prediction diverges with increasing monomial power, likely due to the higher order of error propagation.

\begin{figure}[tb]
    \centering
    \includegraphics[width=1\columnwidth]{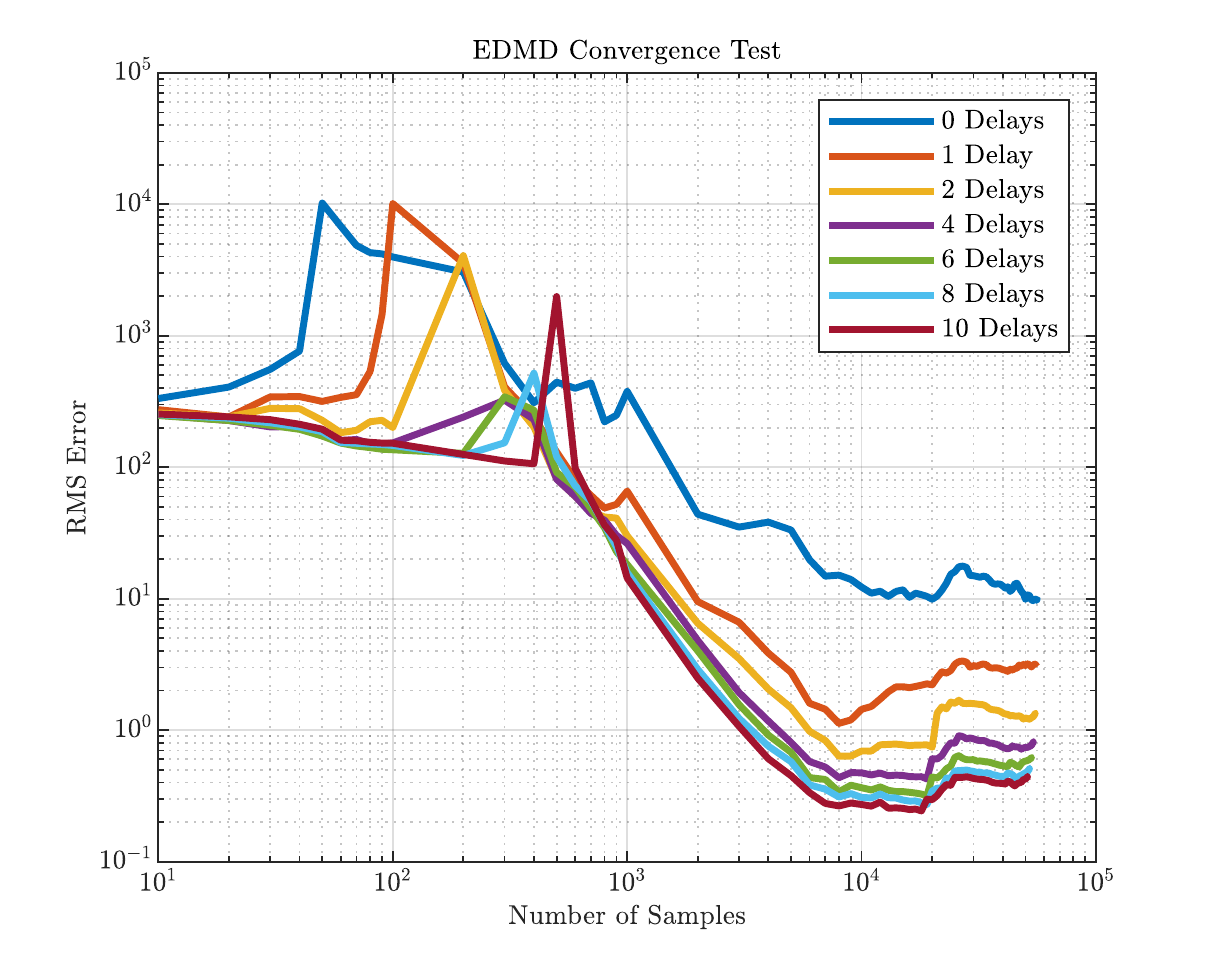}
    \caption{The error $e_{\texttt{RMS}}$ vs number of time samples for a range of time delay observables. The extended state with $i$ delays is formed by stacking the current time sample of $x$ on top of the previous $i$ samples. We see a steady increase in predictive power as the number of time delays increases. Given a certain colored line in this figure, the corresponding monomial-based model with the same number of observables is given by the same color in Fig. \ref{fig:conv_mono}.  For the rest of the paper, we take $10$ time delays.}
    \label{fig:conv_delay}
\end{figure}

Second, we tested time delay observables. Fig. \ref{fig:conv_delay} shows the results of this analysis, with the opposite trend observed compared to Fig. \ref{fig:conv_mono}. We believe this is due to the fact that the momentum of the robot exists in the span of the time delays. We use $10$ time delay observables for the rest of our analysis. 

\subsubsection{Twenty-second Reconstruction}
\begin{figure}[tb]
    \centering
    \includegraphics[width=0.95\columnwidth]{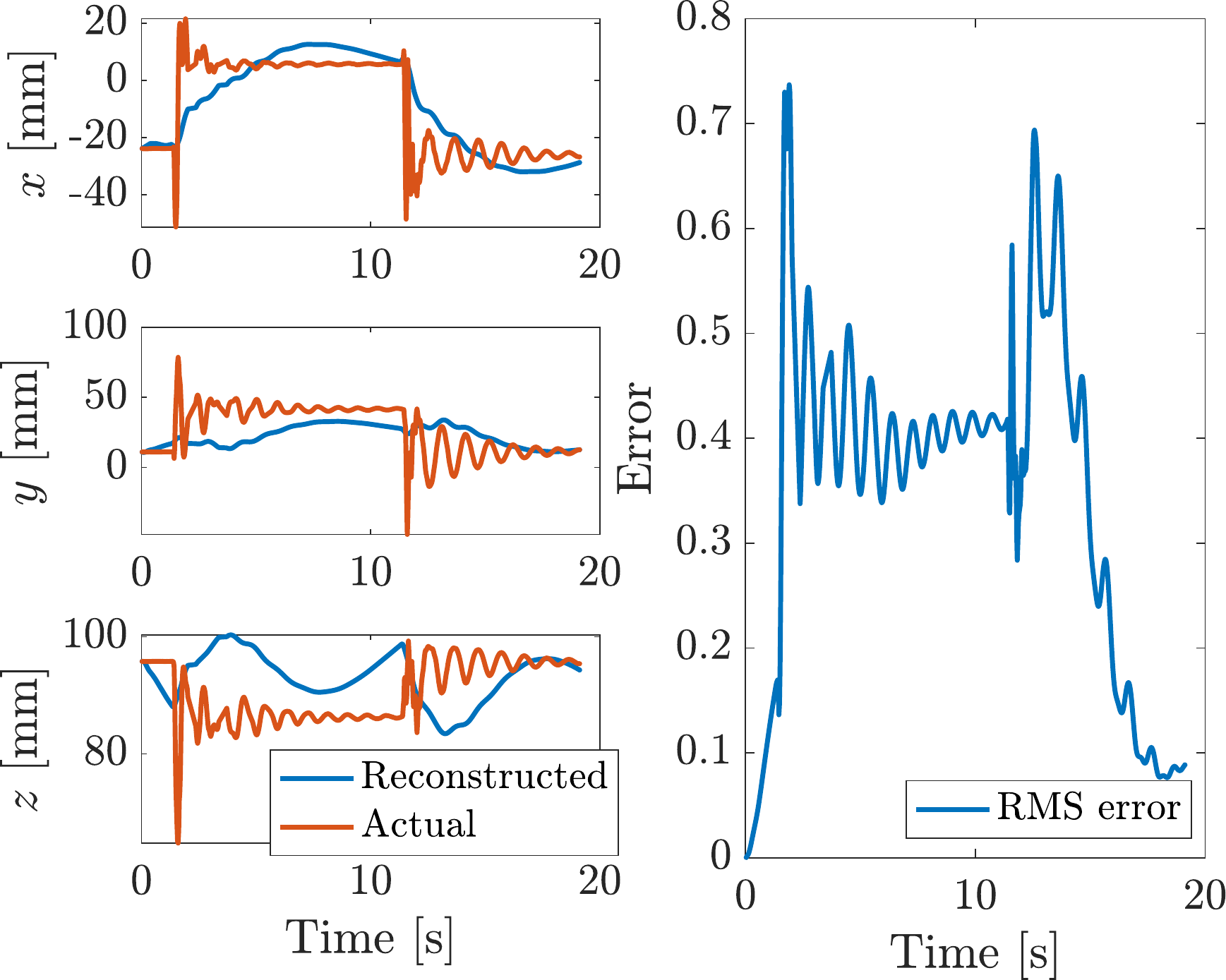}
    \caption{\emph{Left:} Reconstruction of a step input over a 20s time interval, showing the x, y, and z positions (in millimeters). \emph{Right:} Single step RMS reconstruction error.}
    \label{fig:predResults}
    \vspace{-5mm}
\end{figure}

After choosing time delay observables, we attempted to reconstruct a 20 second trajectory. We started at the initial condition for a step input, and iterated forward in time using the known inputs (Fig. \ref{fig:predResults}). While the reconstruction generally tracked the actual behavior, it missed the high frequency ringing found in the real robot. Instead, a lower frequency oscillation appears to be present. That said, given the per-step reconstruction error of ~15\% shown in Fig. \ref{fig:conv_delay}, we note that the tracking error shown in Fig. \ref{fig:predResults} does not propagate as such across hundreds of steps but instead remains bounded.

\subsection{Koopman Optimal Control}
\label{subsec:LQRresults}

After analysing our modeling approach, we then test its efficacy for controlling our system. We first describe our methodology for trimming modes, then show how the reduced order model compares to the full-state model for control.

\subsubsection{Selection of Koopman Modes}
\label{subsubsec:choosingModes}
\begin{figure}[htb]
    \centering
    \includegraphics[width=0.8\columnwidth]{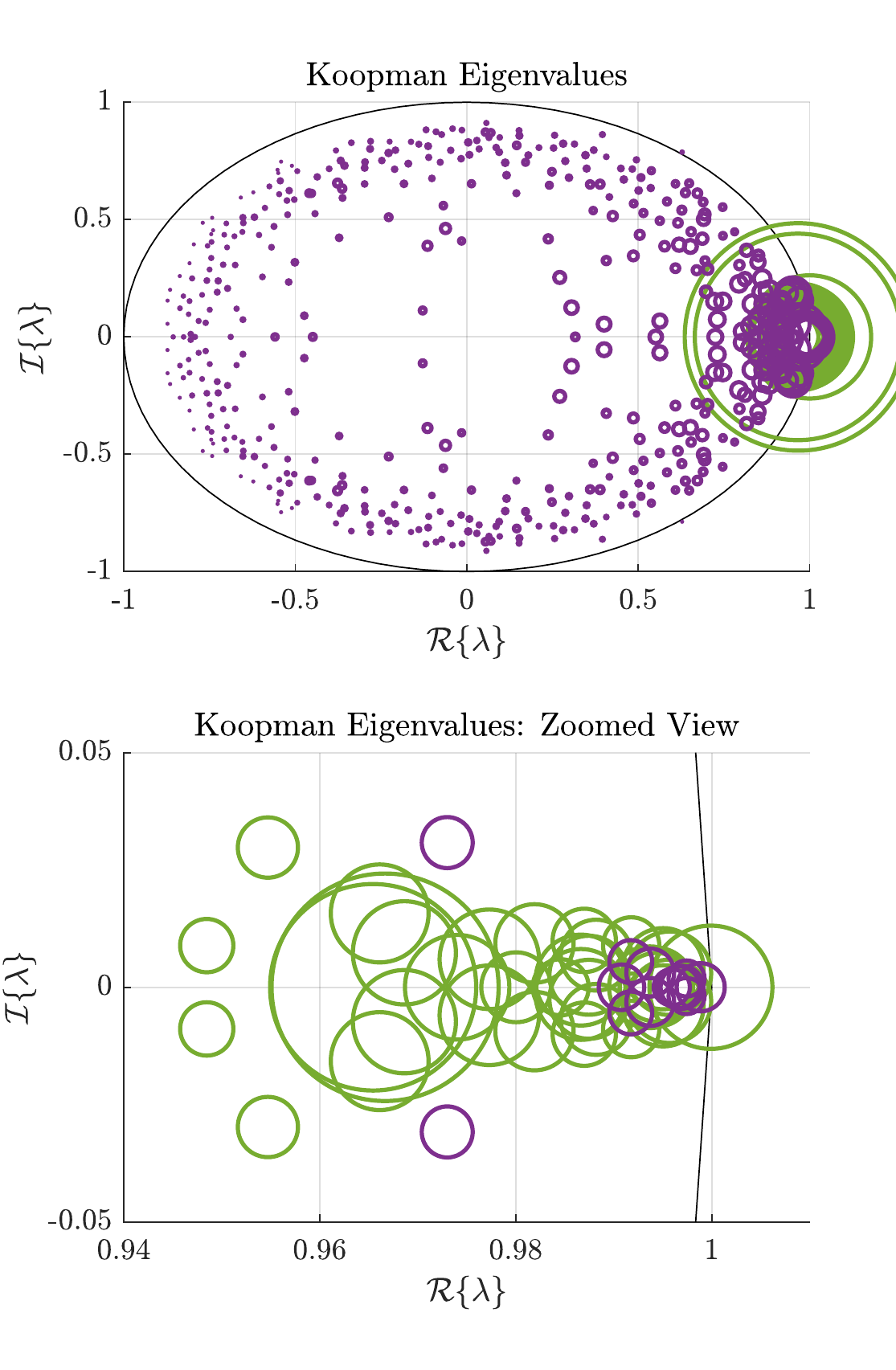}
    \vspace{-5mm}
    \caption{Koopman eigenvalues $\lambda_i$ scaled by their mode powers $|\phi_i(z)|$ averaged over all of the training data. \emph{Top:} The entire distribution of Koopman eigenvalues. \emph{Bottom:} Zoomed in view of the distribution of eigenvalues. The $35$ eigenvalues with the largest mode power are given in green.}
    \label{fig:eig}
    \vspace{-2mm}
\end{figure}

Fig. \ref{fig:eig} shows the eigenvalues $\lambda_i$ of the resulting HDMD matrix. These eigenvalues approximate the eigenvalues of the Koopman operator and are shown with bubble sizes scaled with respect to their respective approximate Koopman mode powers $|\phi_i(\bold{x})|$ which are evaluated for every $x$ in the training data and averaged. The eigenvalues corresponding to low mode power often correspond to modes associated with measurement noise. Often, these modes can be removed from the model with the added bonus of reducing the dimension of the model. To do this, we build a matrix whose column vectors are the $N$ Koopman modes we wish to keep $V=[\bold{v}_1 \ ... \ \bold{v}_N]$. We then project our state space matrices onto the basis of Koopman modes $\tilde{A}=V^{-1}AV$ and $\tilde{B}=V^{-1}B$. The green eigenvalues in Fig. \ref{fig:eig} represent the $35$ modes with the largest mode power. The corresponding reduced order model produced the controller that successfully executed the static reference tracking problem in the bottom part of Fig \ref{fig:introFig}.


\subsubsection{Control with Reduced Dimension Model}
We choose two poses that the robot has achieved in the training period. We then solve the LQR problem, given in Sec. \ref{subsec:koopmanLQR}, with the full state model and with reduced dimension models, created according to the description in Sec. \ref{subsubsec:choosingModes}. The input sequences are then determined in simulation according to \eqref{eqn:optControl}, and deployed to the robot.

\begin{figure}[ht]
    \centering
    \includegraphics[width=0.95\columnwidth]{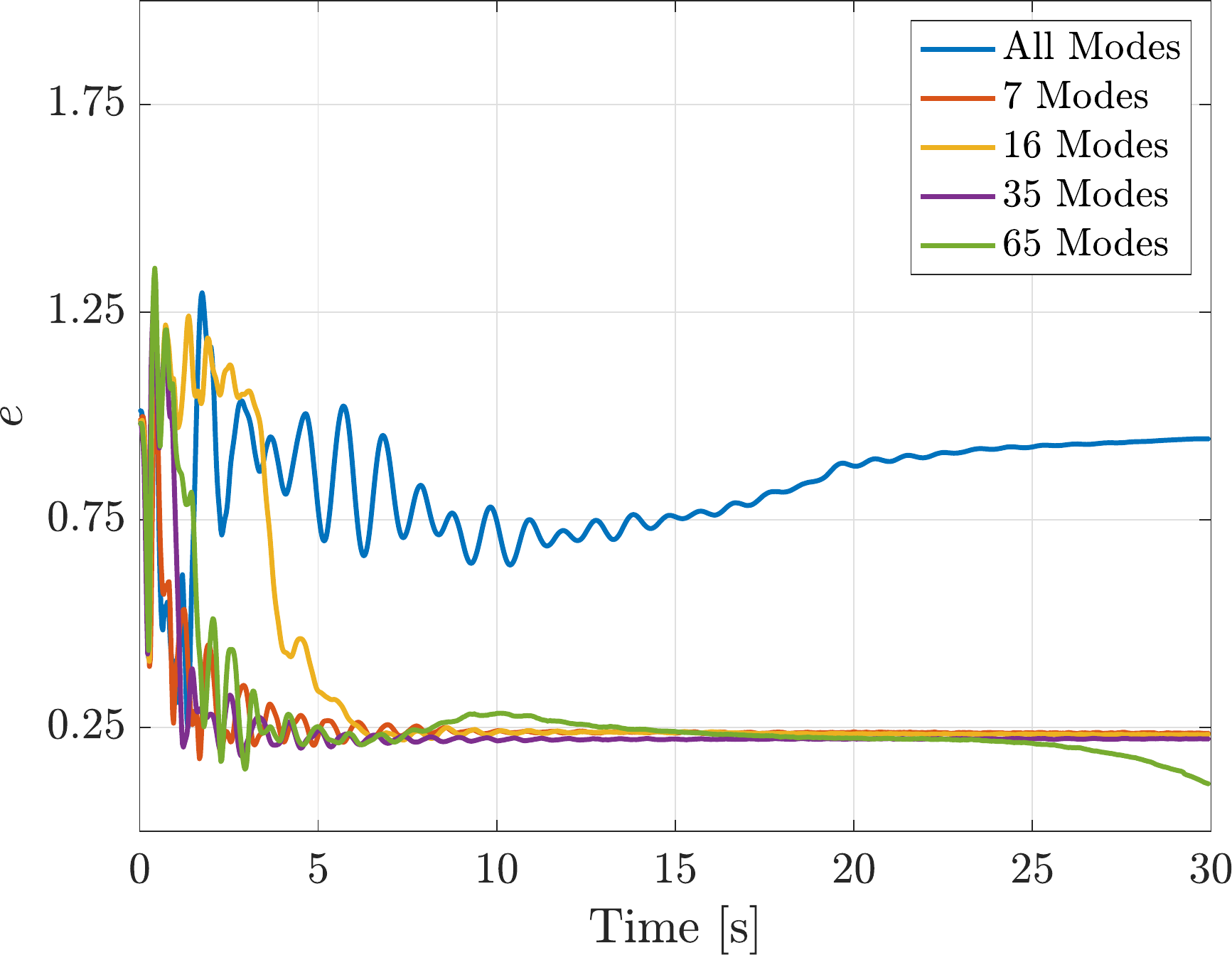}
    \caption{Error $e_{\texttt{RMS}}$ in commanding the robot in open loop to the first pose (shown in Fig. \ref{fig:introFig}) for a variety of mode trimmings.}
    \label{fig:staticLQRpose}
    \vspace{-3mm}
\end{figure}

Figure \ref{fig:staticLQRpose} shows the result for a variety of mode powers for achieving the first pose. This pose is shown in the bottom half of Fig. \ref{fig:introFig}, for the $n=16$ test. In open loop, we are capable of achieving state RMS error of 25\%. We note the full state solution fails to perform as well as the trimmed modes, possibly due to noise present in the full state model.

\begin{figure}[htb]
    \centering
    \includegraphics[width=0.95\columnwidth]{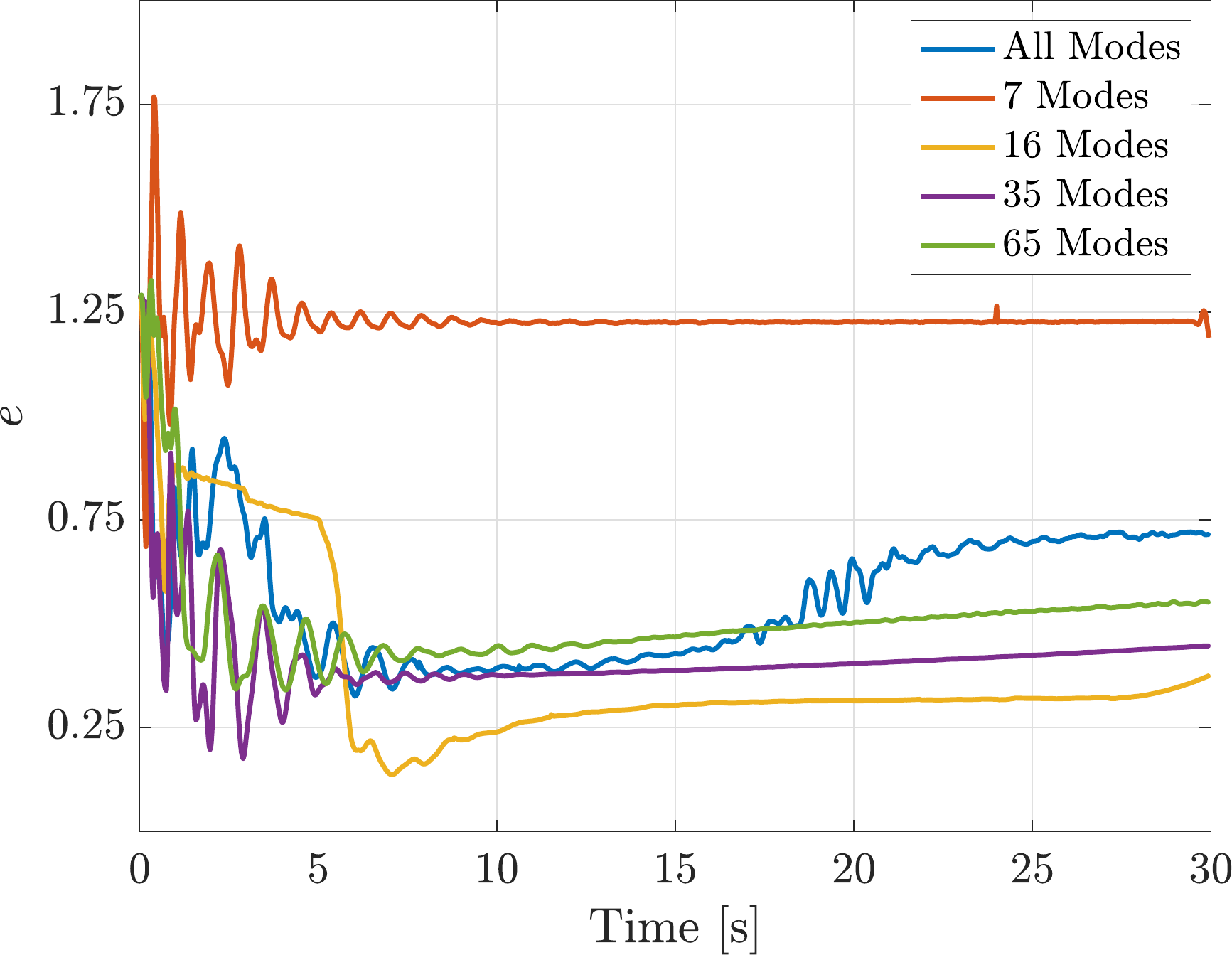}
    \caption{Error $e_{\texttt{RMS}}$ in commanding the robot in open loop to the second pose (shown in Fig. \ref{fig:3dHardPose}) for a variety of mode trimmings.}
    \label{fig:transientLQRpose}
    \vspace{-3mm}
\end{figure}

\begin{figure}[h]
    \centering
    \includegraphics[width = 0.85\columnwidth]{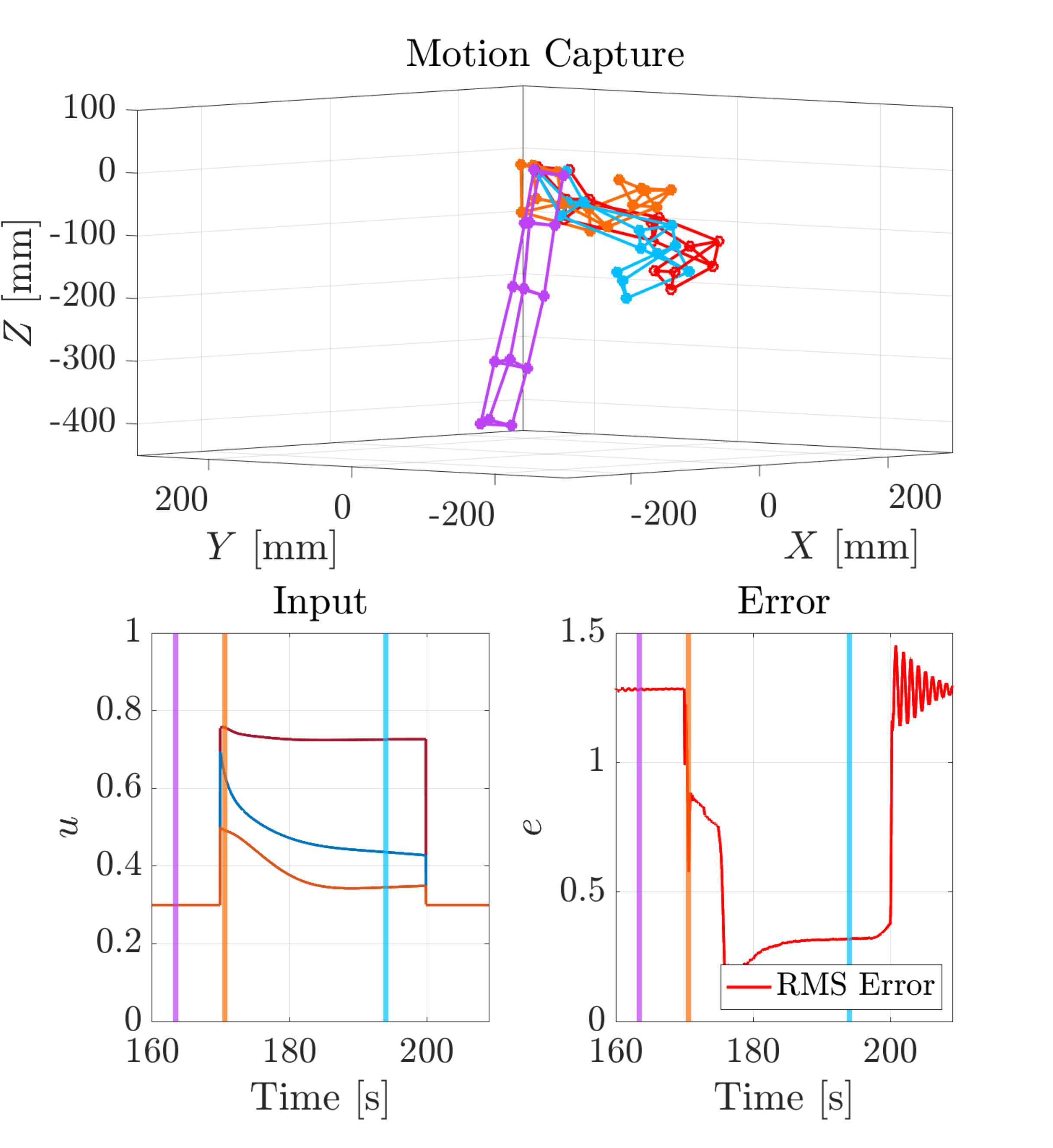}
    \caption{Position, input, and error plots of controlling the robot to the second pose using $n=16$ modes, corresponding to the orange curve in Fig. \ref{fig:transientLQRpose}.}
    \label{fig:3dHardPose}
    \vspace{-5mm}
\end{figure}

We then repeat this process for the second pose (Figs. \ref{fig:transientLQRpose} and \ref{fig:3dHardPose}). 
In this case, all but the 7-mode controller outperform the full state model at steady state (beyond 25 s), although to a lesser extent than in the first case. These two examples suggests that the most prominent dynamics of this robot are effectively captured in only a few Koopman modes. The differential performance across tests for a given set of modes implies that their power is trajectory dependent. If one knows the set of behaviors they desire a robot to complete, this result allows for the informed selection of modes to balance performance and model size.

\section{Discussion and Conclusion}

The potential of soft robots is exciting, with their ability to conform and adapt to unknown environments and embody inherent human safety. Of course, to successfully realize this potential, effective modeling and control approaches must be developed. Here we show that Koopman Operator approaches are a viable path of investigation towards these ends. We present an approach to produce an approximate, low order model with relatively little data, minimal computational cost, and no \emph{a priori} understanding of the input-output dynamics of the system. This approach is also amenable to traditional linear control schemes, such that existing strategies can produce viable control laws, even in open loop, for these highly nonlinear, inertial, and underactuated systems. 

That said, this work is still preliminary. We do not yet capture the fast dynamics of this system, and see our model predicting significantly slower evolution than the real robot achieves. Additionally, we have limited our study to open loop control to understand the extent to which our model can predict behavior, but without closed-loop control, we are limited in the tasks we can command this robot to complete. Finally, much work is to be done in understanding the optimal, minimal selection of modes to achieve the desired behavior. Future work is planned to both advance the construction of our models to simultaneously reject noise and capture faster dynamics, and to implement closed loop control to understand the edges of behaviors this robot can be commanded to display.


\addtolength{\textheight}{-9cm}   






\bibliographystyle{IEEEtran}
\bibliography{IEEEabrv,references}

\begin{thebibliography}{10}
\providecommand{\url}[1]{#1}
\csname url@samestyle\endcsname
\providecommand{\newblock}{\relax}
\providecommand{\bibinfo}[2]{#2}
\providecommand{\BIBentrySTDinterwordspacing}{\spaceskip=0pt\relax}
\providecommand{\BIBentryALTinterwordstretchfactor}{4}
\providecommand{\BIBentryALTinterwordspacing}{\spaceskip=\fontdimen2\font plus
\BIBentryALTinterwordstretchfactor\fontdimen3\font minus
  \fontdimen4\font\relax}
\providecommand{\BIBforeignlanguage}[2]{{%
\expandafter\ifx\csname l@#1\endcsname\relax
\typeout{** WARNING: IEEEtran.bst: No hyphenation pattern has been}%
\typeout{** loaded for the language `#1'. Using the pattern for}%
\typeout{** the default language instead.}%
\else
\language=\csname l@#1\endcsname
\fi
#2}}
\providecommand{\BIBdecl}{\relax}
\BIBdecl

\bibitem{rus2015design}
D.~Rus and M.~T. Tolley, ``Design, fabrication and control of soft robots,''
  \emph{Nature}, vol. 521, no. 7553, pp. 467--475, 2015.

\bibitem{laschi2016soft}
C.~Laschi, B.~Mazzolai, and M.~Cianchetti, ``Soft robotics: Technologies and
  systems pushing the boundaries of robot abilities,'' \emph{Science Robotics},
  vol.~1, no.~1, p. eaah3690, 2016.

\bibitem{trivedi2008soft}
D.~Trivedi, C.~D. Rahn, W.~M. Kier, and I.~D. Walker, ``Soft robotics:
  Biological inspiration, state of the art, and future research,''
  \emph{Applied bionics and biomechanics}, vol.~5, no.~3, pp. 99--117, 2008.

\bibitem{hannan2003kinematics}
M.~W. Hannan and I.~D. Walker, ``Kinematics and the implementation of an
  elephant's trunk manipulator and other continuum style robots,''
  \emph{Journal of robotic systems}, vol.~20, no.~2, pp. 45--63, 2003.

\bibitem{trivedi2008geometrically}
D.~Trivedi, A.~Lotfi, and C.~D. Rahn, ``Geometrically exact models for soft
  robotic manipulators,'' \emph{IEEE Transactions on Robotics}, vol.~24, no.~4,
  pp. 773--780, 2008.

\bibitem{jones2006practical}
B.~A. Jones and I.~D. Walker, ``Practical kinematics for real-time
  implementation of continuum robots,'' \emph{IEEE Transactions on Robotics},
  vol.~22, no.~6, pp. 1087--1099, 2006.

\bibitem{wang2020dexterous}
S.~Wang, R.~Zhang, D.~A. Haggerty, N.~D. Naclerio, and E.~W. Hawkes, ``A
  dexterous tip-extending robot with variable-length shape-locking,''
  \emph{arXiv preprint arXiv:2003.09113}, 2020.

\bibitem{greer2018soft}
J.~D. Greer, T.~K. Morimoto, A.~M. Okamura, and E.~W. Hawkes, ``A soft,
  steerable continuum robot that grows via tip extension,'' \emph{Soft
  robotics}, 2018.

\bibitem{selvaggio2020obstacle}
M.~Selvaggio, L.~Ramirez, N.~Naclerio, B.~Siciliano, and E.~Hawkes, ``An
  obstacle-interaction planning method for navigation of actuated vine
  robots,'' in \emph{2020 IEEE International Conference on Robotics and
  Automation (ICRA)}.\hskip 1em plus 0.5em minus 0.4em\relax IEEE, 2020, pp.
  3227--3233.

\bibitem{falkenhahn2015model}
V.~Falkenhahn, A.~Hildebrandt, R.~Neumann, and O.~Sawodny, ``Model-based
  feedforward position control of constant curvature continuum robots using
  feedback linearization,'' in \emph{2015 IEEE International Conference on
  Robotics and Automation (ICRA)}.\hskip 1em plus 0.5em minus 0.4em\relax IEEE,
  2015, pp. 762--767.

\bibitem{kapadia2010model}
A.~D. Kapadia, I.~D. Walker, D.~M. Dawson, and E.~Tatlicioglu, ``A model-based
  sliding mode controller for extensible continuum robots,'' in
  \emph{Proceedings of the 9th WSEAS international conference on Signal
  processing, robotics and automation}, 2010, pp. 113--120.

\bibitem{marchese2016dynamics}
A.~D. Marchese, R.~Tedrake, and D.~Rus, ``Dynamics and trajectory optimization
  for a soft spatial fluidic elastomer manipulator,'' \emph{The International
  Journal of Robotics Research}, vol.~35, no.~8, pp. 1000--1019, 2016.

\bibitem{engel2006learning}
Y.~Engel, P.~Szabo, and D.~Volkinshtein, ``Learning to control an octopus arm
  with gaussian process temporal difference methods,'' in \emph{Advances in
  neural information processing systems}, 2006, pp. 347--354.

\bibitem{braganza2007neural}
D.~Braganza, D.~M. Dawson, I.~D. Walker, and N.~Nath, ``A neural network
  controller for continuum robots,'' \emph{IEEE transactions on robotics},
  vol.~23, no.~6, pp. 1270--1277, 2007.

\bibitem{thuruthel2017learning}
T.~G. Thuruthel, E.~Falotico, F.~Renda, and C.~Laschi, ``Learning dynamic
  models for open loop predictive control of soft robotic manipulators,''
  \emph{Bioinspiration \& biomimetics}, vol.~12, no.~6, p. 066003, 2017.

\bibitem{della2020data}
C.~Della~Santina, R.~L. Truby, and D.~Rus, ``Data--driven disturbance observers
  for estimating external forces on soft robots,'' \emph{IEEE Robotics and
  Automation Letters}, vol.~5, no.~4, pp. 5717--5724, 2020.

\bibitem{della2020model}
C.~Della~Santina, R.~K. Katzschmann, A.~Bicchi, and D.~Rus, ``Model-based
  dynamic feedback control of a planar soft robot: Trajectory tracking and
  interaction with the environment,'' \emph{The International Journal of
  Robotics Research}, vol.~39, no.~4, pp. 490--513, 2020.

\bibitem{Koopman1931}
B.~O. Koopman, ``Hamiltonian systems and transformation in hilbert space,''
  \emph{Proceedings of the National Academy of Sciences}, vol.~17, no.~5, pp.
  315--318, may 1931.

\bibitem{bruder2019nonlinear}
D.~Bruder, C.~D. Remy, and R.~Vasudevan, ``Nonlinear system identification of
  soft robot dynamics using koopman operator theory,'' in \emph{2019
  International Conference on Robotics and Automation (ICRA)}.\hskip 1em plus
  0.5em minus 0.4em\relax IEEE, 2019, pp. 6244--6250.

\bibitem{arbabi2017ergodic}
H.~Arbabi and I.~Mezic, ``Ergodic theory, dynamic mode decomposition, and
  computation of spectral properties of the koopman operator,'' \emph{SIAM
  Journal on Applied Dynamical Systems}, vol.~16, no.~4, pp. 2096--2126, 2017.

\bibitem{naclerio2020simple}
N.~D. Naclerio and E.~W. Hawkes, ``Simple, low-hysteresis, foldable, fabric
  pneumatic artificial muscle,'' \emph{IEEE Robotics and Automation Letters},
  vol.~5, no.~2, pp. 3406--3413, 2020.

\bibitem{blumenschein2018helical}
L.~H. Blumenschein, N.~S. Usevitch, B.~H. Do, E.~W. Hawkes, and A.~M. Okamura,
  ``Helical actuation on a soft inflated robot body,'' in \emph{2018 IEEE
  International Conference on Soft Robotics (RoboSoft)}.\hskip 1em plus 0.5em
  minus 0.4em\relax IEEE, 2018, pp. 245--252.

\bibitem{Korda2018koopmpc}
M.~Korda and I.~Mezi{\'{c}}, ``Linear predictors for nonlinear dynamical
  systems: Koopman operator meets model predictive control,''
  \emph{Automatica}, vol.~93, pp. 149--160, jul 2018.

\bibitem{arbabi2017computation}
H.~Arbabi and I.~Mezic, ``Computation of transient koopman spectrum using
  hankel-dynamic mode decompoisition,'' \emph{APS}, pp. G1--009, 2017.

\bibitem{Korda2017converge}
M.~Korda and I.~Mezi{\'{c}}, ``On convergence of extended dynamic mode
  decomposition to the koopman operator,'' \emph{Journal of Nonlinear Science},
  vol.~28, no.~2, pp. 687--710, nov 2017.

\bibitem{anderson2007optimal}
B.~D. Anderson and J.~B. Moore, \emph{Optimal control: linear quadratic
  methods}.\hskip 1em plus 0.5em minus 0.4em\relax Courier Corporation, 2007.

\bibitem{bruder2019modeling}
D.~Bruder, B.~Gillespie, C.~D. Remy, and R.~Vasudevan, ``Modeling and control
  of soft robots using the koopman operator and model predictive control,''
  \emph{arXiv preprint arXiv:1902.02827}, 2019.

\end{thebibliography}

\end{document}